\def\year{2019}\relax
\documentclass[letterpaper]{article} 
\usepackage{aaai19}  
\usepackage{times}  
\usepackage{helvet} 
\usepackage{courier}  
\usepackage[hyphens]{url}  
\usepackage{graphicx} 
\usepackage{algorithm}
\usepackage{amsmath,amssymb}
\usepackage{algorithmic}
\usepackage{cleveref}
\usepackage{graphicx}  
\renewcommand{\vec}[1]{\boldsymbol{\mathrm{#1}}}
\newcommand{\mat}[1]{\boldsymbol{\mathrm{#1}}}
\newcommand{\trans}{^{\mkern-1.5mu\mathsf{T}}}

\crefname{figure}{Fig.}{Fig.}
\Crefname{figure}{Fig.}{Fig.}

\title{Adaptable Human Intention and Trajectory Prediction\\for Human-Robot Collaboration}
\author{Abulikemu Abuduweili*,\textsuperscript{\rm 1,3} \ Siyan Li*, \textsuperscript{\rm 2,3}\ and Changliu Liu \textsuperscript{\rm 3}\thanks{The first two authors contributed equally to this paper.}\thanks{This work was supported in part by Holomatic.}\\ 
\textsuperscript{\rm 1}Peking University, Beijing, P. R. China 100871\\
\textsuperscript{\rm 2}Georgia Institute of Technology, Atlanta, GA, USA 30332\\
\textsuperscript{\rm 3}Carnegie Mellon University, Pittsburgh, PA, USA 15213\\
abduwali@pku.edu.cn,\\ 
lisiyansylvia@gatech.edu,\\
cliu6@andrew.cmu.edu\\
}

\begin{document}

\maketitle

\begin{abstract}
To engender safe and efficient human-robot collaboration, it is critical to generate high-fidelity predictions of human behavior. The challenges in making accurate predictions lie in the stochasticity and heterogeneity in human behaviors. This paper introduces a method for human trajectory and intention prediction through a multi-task model that is adaptable across different human subjects. We develop a nonlinear recursive least square parameter adaptation algorithm (NRLS-PAA) to achieve online adaptation. The effectiveness and flexibility of the proposed method has been validated in experiments. In particular, online adaptation can reduce the trajectory prediction error by more than 28\% for a new human subject. The proposed human prediction method has high flexibility, data efficiency, and generalizability, which can support fast integration of HRC systems for user-specified tasks.
\end{abstract}

\section{Introduction}
\noindent 
Recent technological development demands more human-robot collaborations (HRC) in various domains, e.g., industry, service, health care, etc. It is crucial to ensure HRC take place safely and efficiently. 
High-fidelity human behavior prediction is key to safe and efficient HRC. In this paper, we describe algorithms for two tasks: human trajectory prediction and intention prediction. The challenges for building accurate prediction models lie in the following aspects. Firstly, it has been impossible to create a general artificial intelligence to accurately predict human behavior in all circumstances. Encapsulating all human motions into one model is difficult due to their irregular nature. Secondly, human behaviors are highly stochastic, in addition to being heterogeneous. Different motion patterns exist among different people and the same individual repeating a motion. 

To address these problems, this paper proposes a framework to 1) efficiently learn high-fidelity prediction models for user-specified tasks with a small dataset, and 2) ensure adaptability of the model in real time to account for behavioral stochasticity and heterogeneity. This framework supports fast integration of collaborative robots for end users. The pipeline of the framework includes 1) task specification with an and-or graph, 2) human demonstration, 3) learning a multi-task prediction model on demonstrated trajectories, and 4) real-time execution with adaptable prediction enabled by the nonlinear recursive least square parameter adaptation algorithm (NRLS-PAA). With a basic robot collaboration system that handles robot planning and control with respect to the predicted human behavior, an HRC system can be successfully integrated in step 4.

The contributions of this paper include: 1) a framework to support fast integration of HRC systems for end users; 2) a multi-task model for simultaneous trajectory and intention prediction; and 3) NRLS-PAA-based model adaptation to improve online prediction accuracy.

The remainder of this paper will be structured as follows. We will present related works first, followed by the detailed discussion of the proposed framework and method. This paper will then discuss the experiment results, before concluding with future works.

\section{Related Works}
Various HRC systems have been proposed \cite{chandrasekaran2015human}, e.g., robot co-workers in assembly lines \cite{gleeson2013gestures}, surgical robotics \cite{lum2009raven}, and socially interactive robotics helping elderly people \cite{vincze2014towards}. These works all point out the necessity of human behavior prediction.

Current work in human motion prediction tends to separate trajectory and intention predictions. Prediction of human trajectories, especially full arm trajectories or hand trajectories, can rely on either  vision-based methods \cite{Butepage_2017_CVPR} or methods that require wearable devices to monitor neural activities \cite{pistohl2008prediction}. 
Intention can also be predicted using similar wearable sensing systems \cite{Wang:2018:HIP:3173386.3177025}. However, wearable devices may impede the efficiency of HRC systems. Therefore, vision-based methods are more desirable. Various deep neural network-based prediction models that take visual inputs have been proposed. A detailed survey can be found in \cite{DBLP:journals/corr/abs-1905-06113}.

\begin{figure*}[htb]
    \centering
    \resizebox{!}{.4\columnwidth}{
        \includegraphics{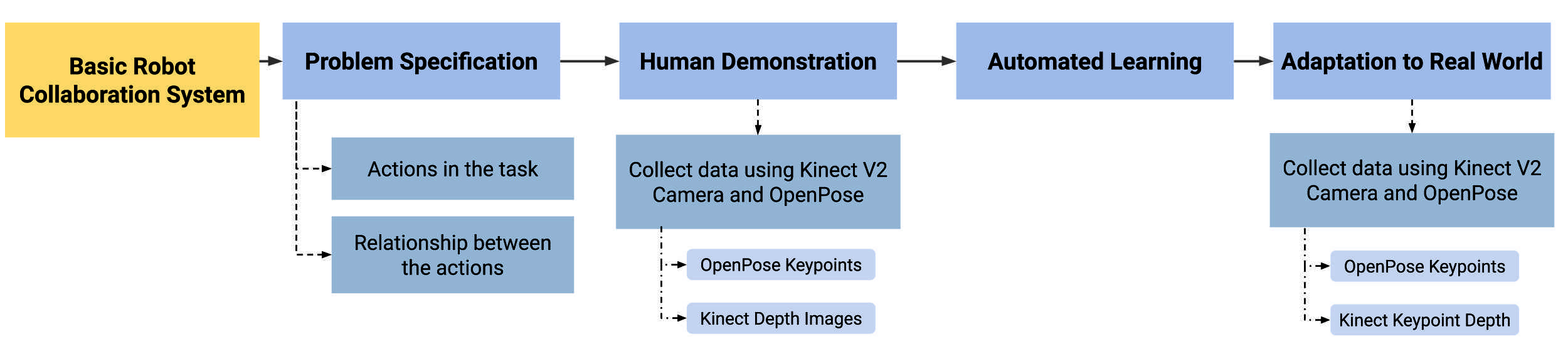}
    }
    \caption{Pipeline for the deployment of a collaborative robot that is able to predict real-time human behaviors.}
    \label{fig:integration}
\vspace{-10pt}
\end{figure*}

To account for individual differences, it is desired for the prediction models to be online adaptable. Existing methods include adapting a feedforward neural network with smooth activation functions using the expectation-maximization (EM) algorithm \cite{ravichandar2016human}, adapting the last layer of a feedforward neural network using the recursive least square parameter adaptation algorithm (RLS-PAA) \cite{Cheng-2019-113166}, and adapting the last layer of a recurrent neural network using RLS-PAA \cite{Si-2019-113165}. 

To improve data efficiency, this paper will propose a multi-task prediction model combining intention and trajectory prediction. The model is more complex compared to the previous models in online adaptation that adapt the last layer, a practice that may not yield desirable results. NRLS-PAA, which is able to handle nonlinear and nonsmooth mappings in our task, will be introduced to adapt the network parameters. 
\section{Methodology}
This section will discuss in detail the framework (the four-step pipeline shown in \cref{fig:integration}) for efficient integration of HRC systems, and the method to learn a multi-task prediction model on small datasets and online adaptation of the model. A card-making task will be used as an example. It is assumed that a basic robot collaboration system as discussed in \cite{Liu-2018-113028} is available to generate intelligent robot control. 

\subsection{Problem Specification}
The tasks learnable by our framework can be represented using and-or graphs \cite{de1990and} and divided up into correlated atomic actions. Currently, the user needs to manually identify both the actions and their respective relationships.
Supposing that the human actions can always be correctly identified and predicted, the and-or graph is able to inform the robot the human's progress based on the prediction and help the robot determine its response.

\begin{figure}[htb]
    \centering
    \resizebox{.9\columnwidth}{!}{
        \includegraphics{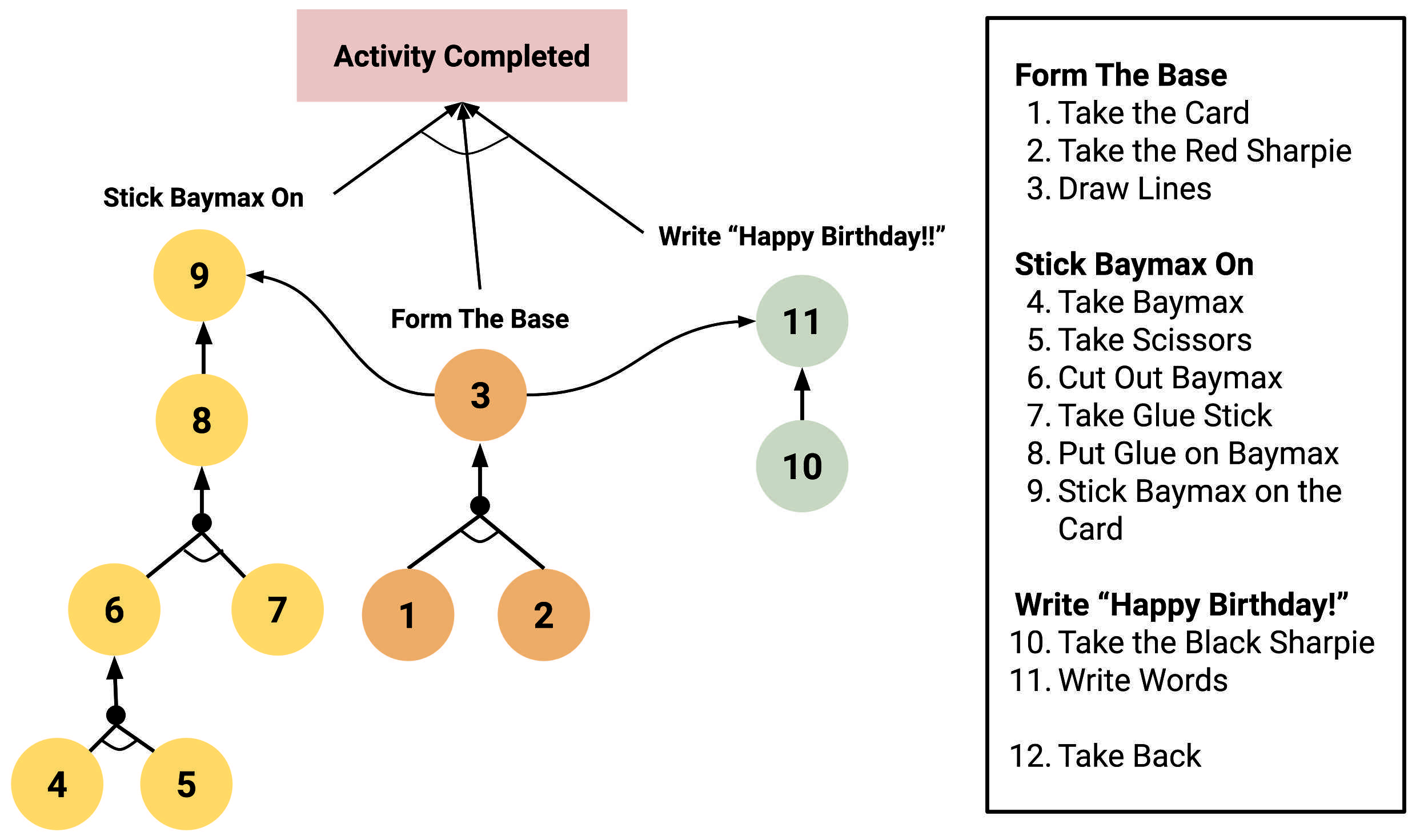}
    }
    \caption{The actions, the relationship between individual actions and their high-level groupings for this task, represented as an and-or graph.}
    \label{fig:action_relations}
\vspace{-10pt}
\end{figure}

We designed a simple card-making task with a goal to make a Baymax birthday card \footnote{The dataset used in this experiment is available at \url{https://github.com/intelligent-control-lab/Intent_Traj_Prediction}.}. The actions and their relationships are illustrated in \cref{fig:action_relations}. 
In this task, one can separate the 12 actions in \cref{fig:action_relations} into three groups with regard to their higher-level purpose, namely ``Form the Base" (scribble lines on the card using the red sharpie), ``Stick Baymax On" (put glue on the back of Baymax after cutting it out, stick it onto the card), and ``Write \textit{Happy Birthday!}" (with the black sharpie). Within each group, every ``reach and take" action must occur first, and the remainder of the actions must occur sequentially. 
We have grouped together the retraction motion from all the take actions (actions 1, 2, 4, 5, 7, and 10) into action 12 for convenience in intention labeling.
In this task, all items are placed in fixed positions so that human intentions can be inferred purely from human trajectories.

\subsection{Human Demonstration}
To prepare for a new collaborative task, the user needs to demonstrate all of its atomic actions to the robot multiple times. Those demonstrated trajectories will be used to train a neural network for predicting human intention and human trajectory in this task.

\subsection{Learning for Trajectory and Intention Prediction}
To create a network suitable for our framework, we construct an encoder-decoder-classifier structure capable of simultaneous intention and trajectory prediction.

\begin{figure}[htb]
    \centering
    \resizebox{.99999 \columnwidth}{!}{
        \includegraphics{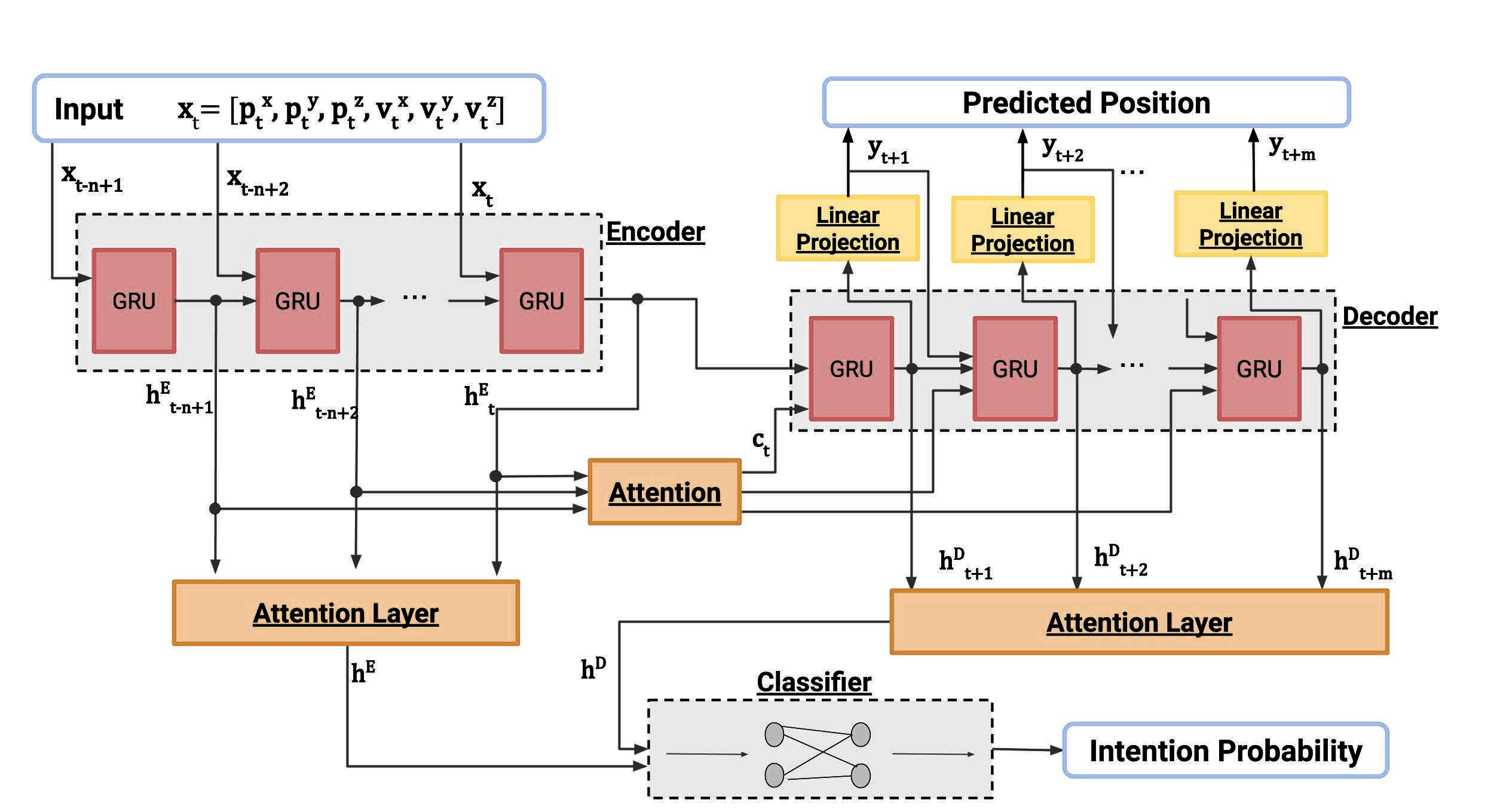}
    }
    \caption{The structure of the neural network in our framework.}
    \label{fig:neural}
\vspace{-10pt}
\end{figure}

\subsubsection{Trajectory Prediction}
Trajectory prediction is based on the encoder-decoder structure of the general sequence-to-sequence (Seq2Seq) model \cite{sutskever2014sequence}.
Both the encoder and the decoder use Gated Recurrent Units (GRU's) \cite{DBLP:journals/corr/ChoMBB14} to generate hidden state vectors. The attention mechanism ~\cite{bahdanau2014neural} is applied to the output vectors of the encoder. 

The input to the neural network are the positions and the velocities of the human joints (only the right wrist for now) at the past $n$ time steps in the x, y, and z directions in Cartesian coordinates. 
The position and velocity sequence is fed into the encoder. 
At time step~$t$, the encoder takes information from $\vec{x_t}$ in addition to the hidden state vector from the previous time step $t - 1$,
\begin{equation}
\vec{h_t^{E}} = {\rm GRU}^{E}( \vec{h_{t - 1}^{E}},\vec{x_t}),
\end{equation}
where $\vec{h_t^{E}}$ represents the encoder's hidden vector at time step $t$, 
and $\vec{x_t}$ the coordinates and velocity at time step~$t$.

Similar to the encoder, the decoder of this network also adopts GRU's. Each decoder unit 
collects the hidden state vector $\vec{h_{t - 1}^{D}}$, a generated output vector $\vec{y_{t - 1}^{D}}$ from the previous time step, and a historical trajectory context vector~$\vec{c_t}$,
\begin{eqnarray}
\vec{h_t^{D}}  &=& {\rm GRU}^{D}( \vec{h_{t - 1}^{D}}, \vec{[y_{t - 1}^{D}};\vec{c_t}]), \\
\vec{c_t}  &=& \sum_{i}{a_{t,i} \cdot \vec{h_i^E}}, \\
a_{t,i}  &=& {\rm score}(\vec{h_{t - 1}^{D}},\vec{h_i^E}),
\end{eqnarray}
where $[\cdot;\cdot]$ denotes vector concatenation.
The context vector $\vec{c_t}$ is a weighted sum of the encoder's hidden states and carries information from the historical trajectory. 
$a_{t,i}$ is a similarity score between the decoder's and the encoder's hidden states \cite{luong2015effective} (e.g. cosine similarity). 
Finally, the decoder maps the hidden state $\vec{h_t^{D}}$  to the output vector $\vec{y_t}$ via a linear projection: $\vec{y_t} = \mat{W^{O}} \cdot \vec{h_t^{D}}$. 

\subsubsection{Intention Prediction}
In our framework, intentions refer to unrealized human actions. Intention prediction uses an encoder-classifier structure. This encoder is designed to be identical to the encoder for trajectory prediction.  

To predict the human intention, our model uses both the hidden vector from the historical trajectory (i.e., from the encoder) and the hidden vector from the predicted future trajectory (i.e., from the decoder). The hidden vectors from both the encoder and the decoder are passed into the attention unit, which produces a weighted sum across different time steps. Finally, these vectors are fed to the fully connected classifier with a softmax activation function, which computes the probability distribution across all intentions,
\begin{equation}
\vec{proba} = {\rm Classifier}([\vec{h^{E}}; \vec{h^{D}}]),
\end{equation}
where $\vec{h^{E}}, \vec{h^{D}}$ are the weighted hidden vectors of the encoder and the decoder, respectively, which are computed by the attention mechanism $\vec{h^{E}}  = \sum_{t}{\mat{W^E} \vec{(h_t^E)\trans} \vec{h_t^E}}$ and $\vec{h^{D}}  = \sum_{t}{ \mat{W^D} \vec{(h_t^D)\trans} \vec{h_t^D}}$,  
where $\mat {W^E}$ and $\mat{W^D}$ are learnable weights.  

\subsubsection{Training Loss Function}
We combine the objective functions for both intention and trajectory prediction into a multi-task model. We use the following loss function to train the model in an end-to-end manner,
\begin{eqnarray}
l  &=& \gamma l^{\rm classification} + (1-\gamma)l^{\rm regression}. 
\end{eqnarray}
$l^{\rm classification}$ is the loss function for intention prediction (e.g., cross-entropy loss), and $l^{\rm regression}$ for trajectory prediction (e.g., L2 loss). Our final loss function is a weighted average of these two loss functions, where the weights are controlled by $ \gamma \in (0,1)$. $ \gamma$ is 0.5 in our experiment.

\subsection{Adapting to Real-world Tasks}
A significant feature of our framework is that it can make adaptable predictions of different individuals. This feature requires the adaptation of parameters in the neural network in real time. 
Adapting arbitrary parameters in a neural network corresponds to a nonlinear least square (NLS) problem: 
Given a data set $\{(\vec{x_i},\vec{y_i}),i=0,1,\cdots\}$, find $\mat{\theta_t} \in {\mathbb{R}}^n$ that minimizes  $J_t(\mat{\theta_t}) = \sum_{i=0}^t{\vec{e_i}^2}$, 
where error term $\vec{e_i} =\vec{y_i}-f(\mat{\theta_t}, \vec{x_i}) $. 
We have developed a recursive algorithm, NRLS-PAA in \cref{code:rekf}, to efficiently solve the NLS problem online. This algorithm is inspired by the recursive Extended Kalman Filter (EKF) method \cite{moriyama2003incremental,alessandri2007recursive}. EKF minimizes the $2$-norm of the state estimation error, which is similar to the NLS problem above. The difference is that: in EKF, the object being estimated is the state value, while in NRLS-PAA, the object is the network parameter. 
NRLS-PAA can adapt arbitrary layers in a network, instead of only the last linear layer with RLS-PAA \cite{Cheng-2019-113166}.

Moreover, a multi-step adaptation strategy is considered in NRLS-PAA. For $k$-step adaptation ($k\in\mathbb{N}$), we stack output and input vectors in $k$ steps respectively as  $\mat{Y_t} = [\vec{y_{t-k+1}},\cdots,\vec{y_t}]$ and $\mat{X_t} = [\vec{x_{t-k+1}},\cdots,\vec{x_t}]$. Assume that each stacked input-output pair is generated via an unknown continuous function $ \mat{Y_t} =  f(\mat{\theta},\mat{X_t})$, with parameter $\mat{\theta}$. Then $\mat{\theta}$ can be adapted online using Algorithm 1. 
In Algorithm 1, $\lambda$ is the forgetting factor; $r$ is related to covariance of approximation error $\vec{e_i}$; $\epsilon$ is related to the Riccati equation in EKF; $\mat{K}$ is the Kalman gain matrix. For each $t = 0, 1, \cdots$, $\mat{\theta_t}$ is an estimate of the (unknown) parameter $\mat{\theta}$. In our experiment, $\mat{\theta}$ corresponds to the weights of the encoder's hidden layers.
\begin{algorithm}
\caption{Non-linear Recursive Least-Squares Parameter Adaptation Algorithm (NRLS-PAA).}\label{code:rekf}
\begin{algorithmic}[1]
\REQUIRE{The initialized parameter $\mat{\theta_0}$;}
\ENSURE{Adapted parameter $\mat{\theta_N}$;}
\STATE {Initialize  $p_0 > 0 ;~ \lambda > 0  ;~ r > 0; ~ \epsilon \ge 0 ;$\\}
$\mat{P_0} = p_0 \mat{I}; $\\
\FOR{ $ t = 0, 1, \cdots $}
\STATE  
 $\mat{\hat{Y}_t}=f(\mat{\theta_t},\mat{X_t})$  \\
\STATE
 Get observed value $\mat{Y_t}$ \\
\STATE 
$ \mat{H_t} = \frac{\partial f(\mat{\theta},\mat{X})}{ \partial \theta} \mid_{\mat{\theta}=\mat{\theta_t},\mat{X}=\mat{X_t}} $\
\STATE 
$ \mat{K_t} = \mat{P_t} \cdot \mat{H_t \trans} \cdot (\mat{H_t} \cdot \mat{P_t} \cdot \mat{H_t \trans} + r \mat{I})^{-1} 
$\
\STATE 
$ \mat{P_{t+1}} =\frac{1}{\lambda}(\mat{P_t} - \mat{K_t}\cdot \mat{H_t} \cdot \mat{P_t} + \epsilon \mat{I}) $\
\STATE 
$\mat{\theta_{t+1}} =\mat{\theta_t} + \mat{K_t} \cdot (\mat{Y_t} - \mat{\hat{Y}_t)} $\
\ENDFOR
\end{algorithmic}
\end{algorithm}
\vspace{-10pt}

\section{Experiments}

In the experiment, a Kinect V2 camera was used to capture the trajectories of human subjects' right arms at an approximate frequency of 20Hz. The human subjects were asked to perform the 12 pre-defined actions. 50 trajectories for each action were obtained from actor A (80\% offline training, 20\% offline validation). Another actor B repeated each action 10 times (100\% testing). 
The wrist trajectories from the past $n = 20$ steps were used to predict the wrist trajectories of the future $m = 10$ steps and the intention. The collected data includes: \textbf{(a)} Depth images from the camera; \textbf{(b)} Keypoint positions generated using OpenPose \cite{cao2018openpose} (pixel positions in the colored images from the Kinect camera) for neck, right shoulder, elbow, and wrist. We used a hole-filling filter to smooth the Kinect depth image and a Kalman filter to smooth the trajectories.\footnote{\url{https://github.com/intelligent-control-lab/Kinect_Smoothing}.}

\subsection{Evaluation of the Multi-task Model}
To demonstrate the effectiveness of the multi-task model, we compared the results of our model with the results of the corresponding single-task models.
\subsubsection{Implementation:}
The encoder and decoder were single layers with 64 hidden GRU's. The classifier has two fully connected layers with 64 and 12 units, respectively. We used an Adam optimizer with a 128 batch size and a 0.01 learning rate. The  single-task model for intention prediction corresponds to the isolated encoder-classifier part of our model. The single-task model for trajectory prediction corresponds to the encoder-decoder part of our model. 
\subsubsection{Metrics:}
We used accuracy to evaluate intention prediction and mean-squared-error (MSE, ${\rm cm}^2$ unit) for trajectory prediction. MSE at each time step is computed as,
\begin{equation} 
    {\rm MSE} = \frac{1}{m}\sum_{t=0}^{m-1}{[(y^x_t-\hat{y}^x_t)^2 +(y^y_t-\hat{y}^y_t)^2 + (y^z_t-\hat{y}^z_t)^2] },
\end{equation}
where $y^x, y^y, y^z$ are the Cartesian coordinates for the observed trajectory, and $\hat{y}^x, \hat{y}^y, \hat{y}^z$ the predicted values.
\subsubsection{Results:}
\renewcommand\arraystretch{1.0}
\begin{table}[ht]
\vspace{-8pt}
\centering
\caption{Prediction Results on Multi and Single Task Model.}
\label{tab:multi_tak}
\begin{tabular}{c|c|c}
\hline\hline
              & Accuracy & MSE (cm$^2$)  \\
\hline
(Single) Intention Prediction    & 0.899 & -    \\
\hline
(Single) Trajectory Prediction   & - & 5.909      \\
\hline
(Multi) Our model   & \textbf{0.930}  & \textbf{5.508}       \\
\hline \hline
\end{tabular}
\vspace{-10pt}
\end{table}

Table \ref{tab:multi_tak} shows the results of the single-task and multi-task models. The multi-task model outperforms the intention-prediction and trajectory-prediction single-task models by margins of 3.44\% and 6.79\%, respectively. 
We summarized two potential reasons why the multi-task model outperforms the single-task ones.
Firstly, in multi-task learning, the sub-tasks regulate each other, which prevents over-fitting. 
Secondly, in our multi-task model, the intention classifier's input contains the decoder's hidden states, which forces the model to predict trajectories more accurately to improve intention predictions.

\subsection{Evaluation of the Adaptation Algorithm}

\subsubsection{Implementation:}
We implemented the 1-step, 2-step, and 5-step adaptation strategies. We set $p_0=0.01, ~\lambda=0.999, ~r=0.95, ~\epsilon = 0$ in our experiment.

\subsubsection{Results:}
\renewcommand\arraystretch{1.0}
\begin{table}[t]
\centering
\caption{Prediction Results on NRLS-PAA.}
\label{tab:adaptation}
\begin{tabular}{c|c|c}
\hline\hline 
              & Accuracy & MSE (cm$^2$)  \\
\hline
Without adaptation    & 0.930 & 5.508    \\
\hline
1-step adaptation   & 0.938 & 4.919      \\
\hline
2-step adaptation   & 0.938  & 4.488       \\
\hline
5-step adaptation   & \textbf{0.946}  & \textbf{3.964}       \\
\hline \hline
\end{tabular}
\vspace{-10pt}
\end{table}
Table \ref{tab:adaptation} shows the results of NRLS adaptation. It is clear that NRLS adaptation boosts the performance of both subtasks. For 1-step adaptation, intention accuracy was improved by 0.86\% and trajectory MSE by 10.69\%. 
For the 5-step adaptation, intention accuracy was improved by 1.72\% and trajectory MSE by 28.03\%. 
The running time for the predictions correlates with the sizes of the layers that need to be adapted. For example, running on GeForce RTX 2080 Ti GPU, predictions take an average of 0.03 seconds per sample when adapting the last linear layer of the classifier (192 parameters), but the average time for the encoder's hidden layers (12288 parameters) is 0.3 seconds for per sample.

\section{Conclusion and Discussion}
This paper proposed an adaptable and teachable framework that facilitated human-robot collaborations on user-specified tasks. A multi-task neural network model was learned from human demonstration for simultaneous trajectory and intention prediction of human behaviors. The network outperformed single-task models in both trajectory prediction and intention prediction. NRLS-PAA was used to adapt to different human subjects. Online prediction accuracy was improved by 28\%.


Our framework can be further improved. Although the training is complete, we are yet to integrate the prediction model into the closed-loop robot collaboration system. Future efforts include system integration and further experimentation. Currently, the task-specific and-or-graph needs to be manually defined. We will enable automatic learning of task specifications in the future. Moreover, task-level knowledge (i.e., the and-or graph) can also be incorporated in the prediction model. Another future goal is to enable full arm prediction.

\bibliography{references.bib} 
\bibliographystyle{aaai} 
\end{document}